\documentclass[10pt,twocolumn,letterpaper]{article}
\pdfoutput=1
\usepackage{iccv}
\usepackage{times}
\usepackage{epsfig}
\usepackage{graphicx}
\usepackage{amsmath}
\usepackage{amssymb}
\usepackage{cuted}
\usepackage{capt-of}
\usepackage[skip=0pt]{caption}
\usepackage{subcaption}
\usepackage{booktabs}
\usepackage{multirow}
\usepackage{float}

\usepackage[pagebackref=true,breaklinks=true,letterpaper=true,colorlinks,bookmarks=false]{hyperref}

\iccvfinalcopy 

\begin{document}

\title{Reconstruction of People in Loose Clothing}

\author{Sai Sagar Jinka \and Rohan Chacko \and Astitva Srivastava \and Avinash Sharma \hspace{40pt} P.J. Narayanan \hfill
\\\\
International Institute of Information Technology Hyderabad\\
{\tt\small {\{jinka.sagar, astitva.srivastava\}}@research.iiit.ac.in}\\
{\tt\small {rohan.chacko@students.iiit.ac.in}}\\
{\tt\small {\{asharma, pjn\}}@iiit.ac.in}
}

\maketitle
\begin{strip}
    \centering
    \includegraphics[width=\linewidth]{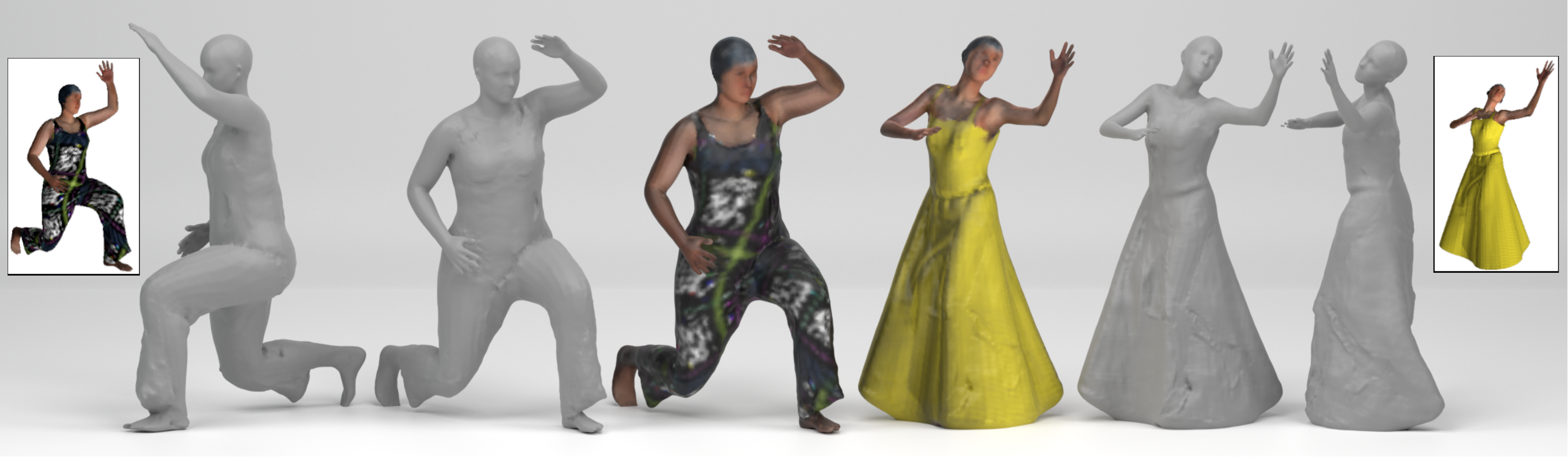}
    \captionof{figure}{
    our method achieves high fidelity textured 3D mesh reconstructions where loose clothing significantly occludes the body.  It successfully recovers the fine-grained geometrical details around cloth surface and plausible geometrical structure for self-occluded body parts. our method is robust to extremely loose clothing, complex poses, and skewed viewpoints.
    \label{fig:teaser}}
\end{strip}

\begin{abstract}
    3D human body reconstruction from monocular images is an interesting and ill-posed problem in computer vision with wider applications in multiple domains. In this paper, we propose a novel end-to-end trainable network that accurately recovers the detailed geometry and appearance of 3D people in loose clothing from a monocular image. We propose a sparse and efficient fusion of a parametric body prior with a non-parametric peeled depth map representation of clothed models.
    The parametric body prior constrains our model in two ways: first, the network retains geometrically consistent body parts that are not occluded by clothing, and second, it provides a body shape context that improves prediction of the peeled depth maps. This enables our method to recover fine-grained 3D geometrical details with just L1 losses on the 2D maps, given an input image.  
    We evaluate our method on publicly available Cloth3D and THuman datasets and report superior performance to state-of-the-art approaches. 
\end{abstract}
\section{Introduction}
    Image-based modeling of loosely clothed 3D humans is an interesting and challenging open problem in computer vision. It has several applications in the domains of entertainment, AR/VR, sports, and healthcare. Traditional solutions based on stereo/multi-view reconstruction ~\cite{gall2009motion,bogo2017dynamic,dou2016fusion4d,guo2019relightables,stoll2011fast} require studio environments with multiple synchronized calibrated cameras, yet have limitations in recovering high-frequency geometrical details.  
    Advancements in commercial depth sensing over the last decade ~\cite{shotton2011real,baak2013data,wei2012accurate,newcombe2015dynamicfusion} has partly helped in overcoming this limitation. Nevertheless, these sensors suffer from motion blur in dynamic scenes. Additionally, both RGB and depth camera-based methods are susceptible to severe self-occlusion artifacts caused by loose clothing and articulated body poses. 
    \\\\
    With the advent of deep learning models, significant interest has garnered around 3D reconstruction from a monocular image~\cite{kanazawa2018end,varol2018bodynet,habermann2020deepcap}, which is an ill-posed problem. Challenges like self-occlusions, viewpoint variations, and clothing obstacles make the scenario more difficult. 
    One class of existing deep learning solutions attempts to fit a parametric body model like SMPL~\cite{SMPL:2015} to a monocular input image using global image features~\cite{kanazawa2018end,guler2018densepose,omran2018neural}.
    Recently, several approaches~\cite{alldieck19cvpr,bhatnagar2019mgn,patel20tailornet,alldieck2019tex2shape} have proposed parametric clothing over the SMPL body. However, these methods can only accommodate relatively tight clothing styles as they primarily estimate local displacement/deformation of original SMPL vertices.
    
    The other class of non-parametric body reconstruction techniques pose no such body prior constraints~\cite{saito2019pifu,saito2020pifuhd,natsume2019siclope,varol2018bodynet,bhatnagar2020loopreg,venkat2018deep} and hence can potentially handle loose clothing scenarios. In particular, the recent implicit function learning models, PIFu~\cite{saito2019pifu} and PIFuHD~\cite{saito2020pifuhd} estimate voxel occupancy by utilizing pixel-aligned RGB image features computed by projecting 3D points onto the input image. However, the pixel-aligned features suffer from depth ambiguity as multiple 3D points are projected to the same pixel. As an alternate representation for 3D objects/scenes, some of the recent works model scenes as multiple (depth) plane images (MPIs)~\cite{single_view_mpi}.  MPI is altogether a layered representation of depths, \ie each layer is at a fixed depth in the camera frustum. 3D human body reconstruction has also been attempted in the same vein by predicting front and back depth maps in~\cite{mouldinghumans}. However, they fail to handle self-occlusions by body parts. 
    Another interesting work~\cite{jinka2020peeledhuman} attempted to address the self-occlusion problem by predicting multiple peeled depth maps. Their sparse representation of the human body was obtained by encoding the 3D body surface using inverse ray-tracing formulation. 
    Nevertheless, these approaches fail to take into account the fine-grain local geometry structure and do not seek to enforce global consistency on the body shape/pose to encourage physically plausible shapes and poses of human body parts. 
    
    The aforementioned problems can be addressed by introducing a body prior while reconstructing humans in loose clothing. 
    The volume-to-volume translation network proposed in~\cite{zheng2019deephuman} attempts to fuse image features with the SMPL prior in a volumetric representation. However, volumetric convolutions are computationally expensive and is limited in resolution.
    Recently, ARCH~\cite{huang2020arch} proposed to induce a human body prior by sampling points around a template SMPL mesh before evaluating occupancy labels for each point.
    However, sampling around the canonical body surface is not sufficient to reconstruct humans with complex articulated poses in loose clothing. 
    Another interesting work~\cite{he2020geo} attempted to refine implicit function estimation by fusing volumetric features and pixel-aligned features to resolve local feature ambiguity. However, volumetric feature estimation is still computationally expensive and their method is not end-to-end trainable.
    Thus, existing methods either lack computational efficiency or compute only small deformations around the SMPL body prior. However, in loose clothing scenarios, garments can also undergo geometrical deformations that are independent of the underlying body shape and pose. Hence, there is an acute need to learn these deformations apart from body shape/pose deformation. This can be achieved by efficiently combining complementary strengths of parametric and non-parametric reconstruction paradigms. 
    
    In this paper, we propose an efficient, sparse, and robust 3D body reconstruction method that can successfully handle significantly loose clothing and skewed viewpoints. Our method combines the SMPL body prior with peeled depth predictions to achieve geometrically rich 3D body reconstruction from a monocular image.
    %
    Starting with a good estimate of the SMPL body from a monocular image ({\emph e.g.}, using~\cite{tan2020self}), 
    we encode this mesh into the sparse 2D peeled depth map representation proposed in~\cite{jinka2020peeledhuman} to obtain SMPL peeled depth maps. 
    The SMPL prior along with the monocular RGB image is fed as input to our network which then predicts a dense additive residual deformation for each pixel in the SMPL peeled depth maps. This additive depth-based deformation accounts for clothing-specific geometrical deformations at each location near the body surface.
    However, such modeling do not focus on reconstructing regions where the clothing is far from the body surface. 
    To reconstruct such regions, we propose to predict peeled depth maps in a separate prediction branch of our network. These predicted peeled maps are later fused with the residual deformation maps along with the initial SMPL peeled depth maps to obtain the final fused peeled depth maps. A separate decoder is used to predict peeled RGB maps. The peeled RGB maps are superimposed on the fused peeled depth maps to recover a colored point-cloud and subsequently a vertex colored mesh. 
    The SMPL prior constrains our model in two ways: first, the network retains the geometrically consistent body parts that are not occluded by clothing, and second, it provides a body shape context that improves prediction of peeled depth maps. This enables our network to learn with only $L_1$ losses unlike other existing methods that use GAN loss and 3D Chamfer loss. 
    %
    \begin{figure*}[t!] 
        \centering
            \includegraphics[width=1\linewidth]{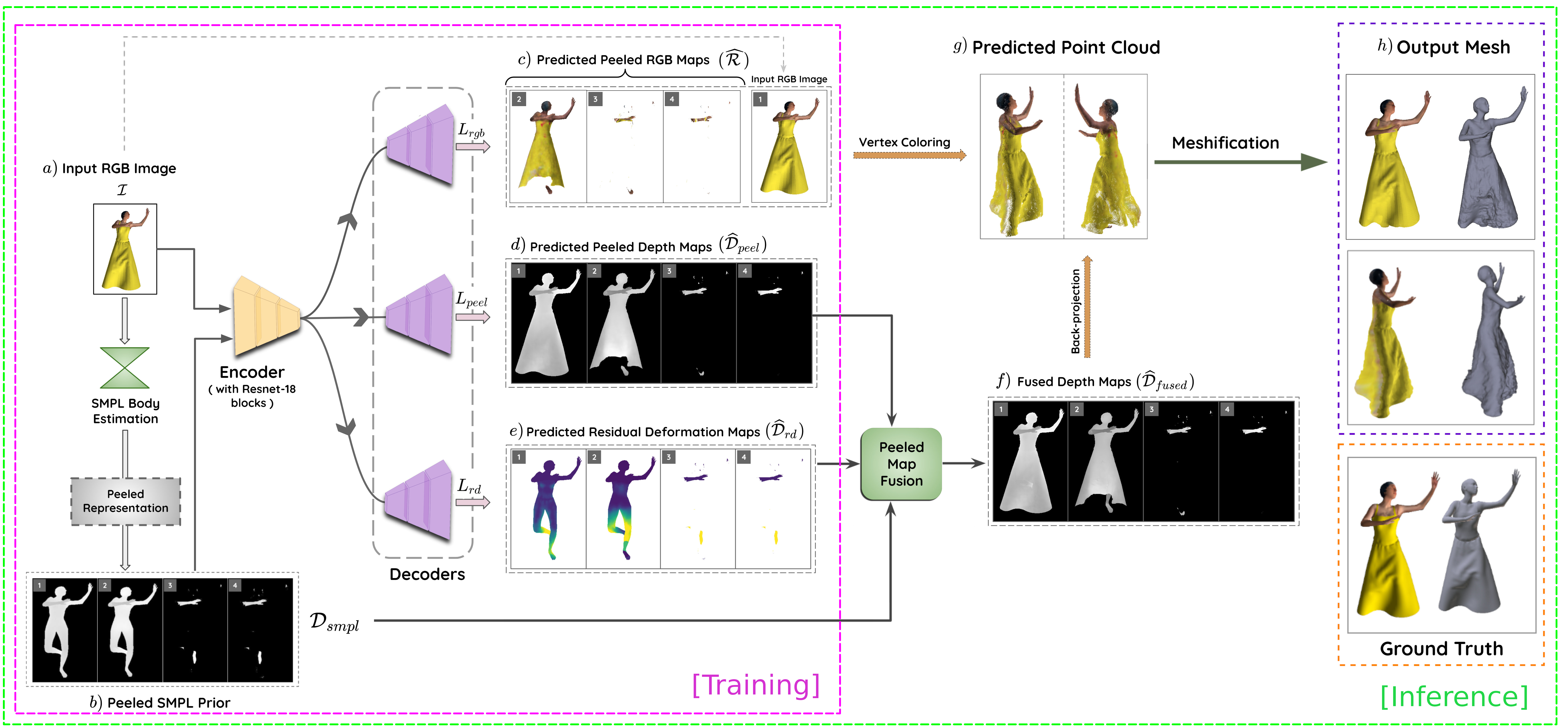}
            \newline
            \caption{\textbf{Pipeline:} We predict an SMPL prior from input image $\mathcal{I}$, and later convert it to peeled depth map ($\mathcal{D}_{smpl}$). This along with  image $\mathcal{I}$, is fed to an encoder. 
            Subsequently, three separate decoders branches predict peeled RGB ($\widehat{\mathcal{R}}$), peeled depth ($\widehat{\mathcal{D}}_{peel}$) and Residual Deformation ($\widehat{\mathcal{D}}_{RD}$) maps, respectively. Finally, a layer-wise fusion of $\widehat{\mathcal{D}}_{peel}$, $\widehat{\mathcal{D}}_{rd}$ and $\mathcal{D}_{smpl}$ is performed to obtain $\widehat{\mathcal{D}}_{fused}$, which is then back-projected along with $\widehat{\mathcal{R}}$ to obtain a per-vertex colored point-cloud. (The Ground Truth mesh is shown for comparison only.)} 
    \label{fig:Pipeline}
    \end{figure*}
    
    Our technical contributions are listed below:
    \begin{itemize}
    \item {We propose a novel end-to-end trainable encoder-decoder architecture our method, that uses only $L1$ losses on 2D maps for predicting a 3D body model with loose clothing from monocular image. 
    }
        \item {Our method achieves sparse and efficient fusion of a parametric body prior with non-parametric peeled depth representation.}
  
        \item{We evaluate our method on various publicly available datasets and report superior qualitative \& quantitative results as compared to state-of-the-art methods.}
    \end{itemize}
\section{Related Work}
    \textbf{Parametric Body Fitting.} 
    Estimating the 3D human body pose~\cite{mehta2017vnect,newell2016stacked} using deep neural networks has achieved great success with robust performance. The naked animatable 3D body is represented by SMPL~\cite{SMPL:2015}, SMPL-X~\cite{pavlakos2019expressive}, SCAPE~\cite{anguelov2005scape}. These models can be estimated from a single image by directly regressing the 3D joint locations. HMR~\cite{kanazawa2018end} proposes to regress SMPL parameters when minimizing re-projection loss with the known 2D joints.  
    Different priors have been used to refine the parametric estimates as in~\cite{varol2017learning,omran2018neural,kolotouros2019learning,kanazawa2019learning}.
    Despite these approaches being computationally efficient, they lack person-specific details. SMPL vertex offset estimates have been proposed to understand tight clothing details~\cite{bhatnagar2019mgn,venkat2019humanmeshnet,kolotouros2019convolutional,Venkat_2019_ICCV_Workshops}. \\
    \\
    \textbf{Non-parametric Body Reconstruction.} Recovering 3D human body from multi-camera setups employ voxel carving, triangulation, multi-view stereo, shape-from-X techniques~\cite{azevedo20093d,dou2016fusion4d,bogo2017dynamic,mulayim2003silhouette}. 
    With the advent of deep learning, initially voxel methods gained popularity as voxels are a natural extension to 2D pixels~\cite{venkat2018deep,varol2018bodynet,zheng2019deephuman}. 
    SiCloPe~\cite{natsume2019siclope} estimates silhouettes in novel views to recover underlying 3D shape. Recent implicit function learning methods for human body reconstruction use 
    locally-aligned pixel features 
    space~\cite{saito2019pifu,saito2020pifuhd}. However, they suffer from sampling limitation and do not model explicit shape representation. 
    Peeled maps~\cite{jinka2020peeledhuman} proposed a sparse representation by estimating only surface intersections by posing the problem as an extension to ray tracing. Similar idea is used for view synthesis by NERF~\cite{mildenhall2020nerf} where it samples points along the camera ray to evaluate RGB$\sigma$ on these samples.\\
    \\
    \textbf{Prior-based Non-Parametric Body Reconstruction.} ARCH~\cite{huang2020arch} combines the SMPL parametric body model with implicit functions to assign skinning weights for the reconstructed mesh. 
    However, the method cannot reconstruct loose clothing in complex poses. Geo-PIFu~\cite{he2020geo} learns deep implicit functions by utilizing structure-aware 3D voxel features along with 2D pixel features. 
    DeepHuman~\cite{zheng2019deephuman} leverages dense semantic representations from SMPL as an additional input. Nevertheless, both Geo-PIFu and DeepHuman are volumetric-regression based approaches that incur high computational costs.

\section{Method}
    
    In this section, we first outline peeled representations for 3D shapes followed by the details of our proposed method.
    
    \subsection{Peeled Representation}
    Peeled representation is a sparse, non-parametric encoding of 3D shapes~\cite{jinka2020peeledhuman}. Each textured 3D body is represented as a set of 2D maps - four depth maps and four RGB maps. The depth and color values are recorded at each intersection of a camera ray with the 3D surface. This representation is more efficient than voxels and implicit functions as it only stores ray-surface intersection in a 2D multi-layered layout.
    Although this representation is able to handle severe self-occlusions, it lacks the ability to predict plausible human body shapes for complex poses as there is no inductive bias on the human body structure. 
    %
    %
    \subsection{Method Overview} 
    \label{subsec:peeled++}
        Given an image $\mathcal{I}$ of a person in arbitrary pose with loose clothing, we aim to reconstruct a textured 3D body surface with high-frequency geometrical details in clothing and plausible body part structure, as shown in \autoref{fig:Pipeline}. There are three main components to our method i) An input peeled SMPL prior ii) Additive Residual Deformation (RD) prediction, and iii) Peeled map fusion. 
        First, we get the SMPL body mesh fitted to the input image and convert it to a peeled depth representation to obtain a peeled shape prior $\mathcal{D}_{smpl}$ as shown in \autoref{fig:Pipeline}(b). This along with $\mathcal{I}$ is fed as input to the shared encoder in our network. The network then predicts three outputs through different decoder branches, namely, peeled RGB maps $\widehat{\mathcal{R}}$, peeled depth maps $\widehat{\mathcal{D}}_{peel}$ and peeled residual deformations $\widehat{\mathcal{D}}_{rd}$, as shown in \autoref{fig:Pipeline}(c-e). The topmost decoder branch predicts only three peeled RGB maps as the input $\mathcal{I}$ naturally acts as the first peeled RGB map. The predicted peeled depth maps $\widehat{\mathcal{D}}_{peel}$ and residual deformation maps $\widehat{\mathcal{D}}_{rd}$ are subsequently combined using $\mathcal{D}_{smpl}$ in a peeled map fusion module to get the final fused peeled depth maps $\widehat{\mathcal{D}}_{fused}$ shown in \autoref{fig:Pipeline}(f). 
        Subsequently, a colored point-cloud representation is obtained by back-projecting $\widehat{\mathcal{D}}_{fused}$ and $\widehat{\mathcal{R}}$ to $3$D coordinates using known camera intrinsics, as shown in \autoref{fig:Pipeline}(g). This point-cloud is further post-processed to obtain a mesh reconstruction using Poisson~\cite{kazhdan2006poisson} method. A detailed discussion on these network components and training protocol including the loss functions is provided below. 
        
        
    \subsubsection{Peeled Shape Prior} 
    \label{subsubsec:geometric_space.}
        We estimate the SMPL model parameters ($\beta$, $\theta$ and global translation/rotation) for an input monocular image $\mathcal{I}$ using~\cite{tan2020self}. 
        The predicted SMPL mesh is converted to sparse peeled depth map representation defined as: 
        \begin{equation}
            \mathcal{D}_{smpl} = \{(x^i,y^i,d^i) : \forall (x,y) \in \mathcal{I}, i \in \{1,2,3,4\},d \in \mathbb{R}\}.
        \end{equation}
        \begin{figure}
             \centering
            \includegraphics[width=0.7\linewidth]{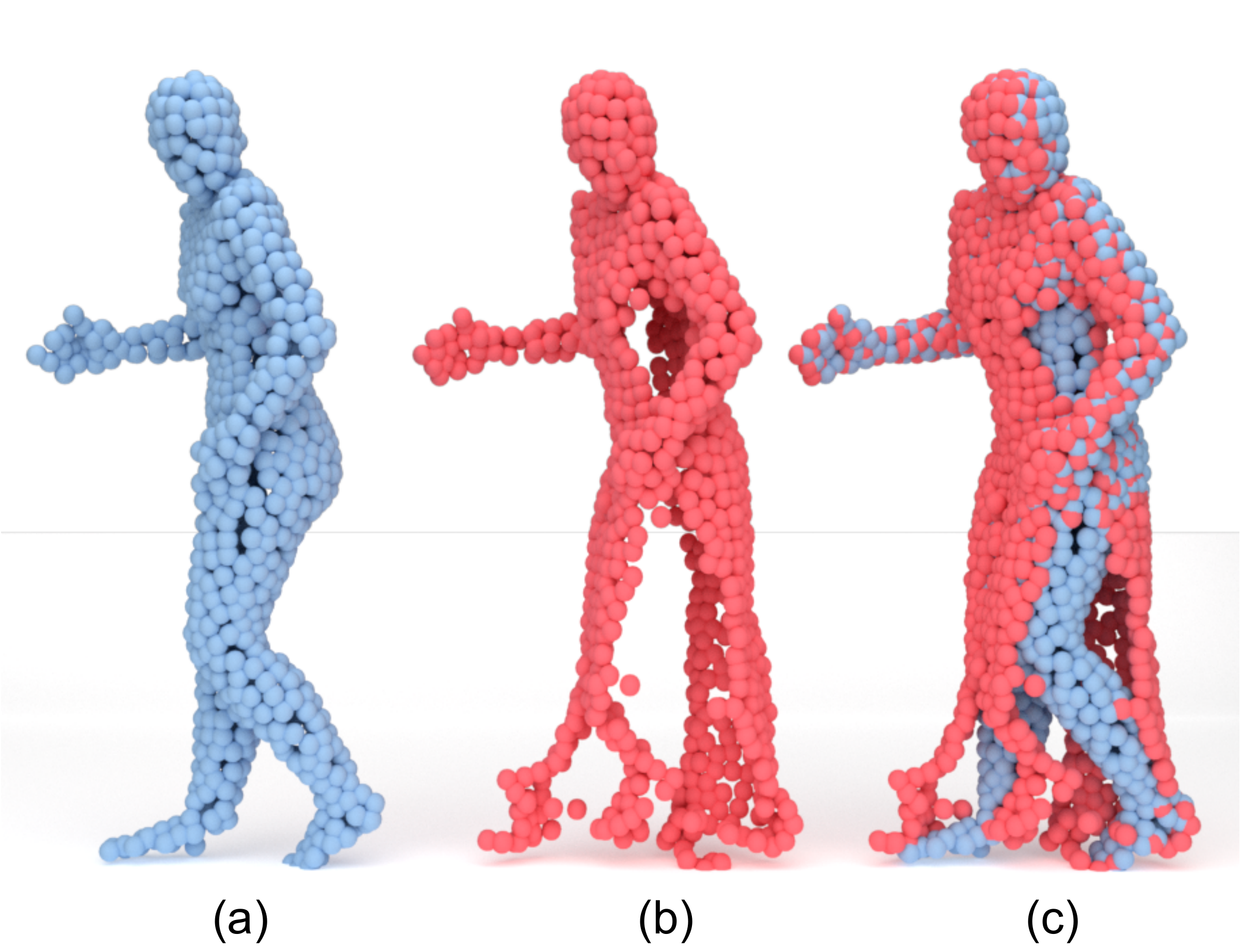}
            \caption{\textbf{Residual Deformation  rendered as point-cloud.} (a) Input ($\mathcal{D}_{smpl}$) shown in blue, (b) Predicted $\widehat{\mathcal{D}}_{rd}$ shown in red. (c) Superimposing $\widehat{\mathcal{D}}_{rd}$ and $\mathcal{D}_{smpl}$.}
            \label{fig:RD}
        \end{figure}
        The input image concatenated with the SMPL peeled depth maps $\mathcal{D}_{smpl}$ forms the input to our network .
    \subsubsection{Residual Deformation (RD)} 
    \label{subsubsec:RD}
        To estimate image specific deformations from the naked SMPL prior input, we propose to predict Residual Deformation maps $\widehat{\mathcal{D}}_{rd}$ by computing pixel-wise offsets from the SMPL peeled depth maps $\mathcal{D}_{smpl}$, defined as:
        \begin{equation}
            \widehat{\mathcal{D}}_{rd} = \{ (x^i,y^i,\widehat{\delta d}^i) : \forall (x,y) \in \mathcal{I}, i \in \{1,2,3,4\},d \in \mathbb{R}\}.
        \end{equation}
        More specifically, this models deformations of the SMPL prior in accordance with the clothing present in the input image. On pixels where there is no clothing (eg. face and hands), it predicts the offsets to be $0$, thereby retaining the geometrical structure of body parts, as shown in \autoref{fig:RD}. 

    \subsubsection{Peeled Map Fusion}\label{subsubsec:fusion}
        
        
        
        The RD map modeling does not focus on regions where loose clothing is far from body surface. 
        Hence, we propose to 
        compute the final fused peeled depth maps $\widehat{\mathcal{D}}_{fused}$ as the fusion of RD maps $\widehat{\mathcal{D}}_{rd}$ and predicted peeled depth map $\widehat{\mathcal{D}}_{peel}$ using the SMPL prior. The $i^{th}$ layer fused peeled map \ie, $\widehat{\mathcal{D}}_{fused}^i \in \widehat{\mathcal{D}}_{fused}$ is defined as:
        \begin{equation}
            \widehat{\mathcal{D}}_{fused}^i = m_{i} \odot (\widehat{\mathcal{D}}_{rd}^i+\mathcal{D}_{smpl}^i) + (1-m_{i}) \odot \widehat{\mathcal{D}}_{peel}^{i}
        \end{equation} \label{eq:fusion}
%
        \[
            m_i= 
        \begin{cases}
            1,& \text{if } \widehat{\mathcal{D}}_{rd}^i > 0 \text{ and } \widehat{\mathcal{D}}_{peel}^{i} > 0\\
            0,              & \text{otherwise}
        \end{cases}
        \]
        where $\odot$ is element-wise multiplication and for each $i^{th}$ layer $\widehat{\mathcal{D}}_{peel}^i \in \widehat{\mathcal{D}}_{peel}$, $\widehat{\mathcal{D}}_{rd}^i \in \widehat{\mathcal{D}}_{rd}$ and $\mathcal{D}_{smpl}^i \in \mathcal{D}_{smpl}$. 
              
        
    \subsection{Training}
        The learning objective is defined as:
        \begin{equation}
            L = L_{peel} + \lambda_{rd} L_{rd} + \lambda_{rgb} L_{rgb} + \lambda_{sm} L_{sm}
        \end{equation}
        where $\lambda_{rd}$, $\lambda_{rgb}$ and $\lambda_{sm}$ are regularization parameters. We provide the formulation for the individual loss terms below.
        \begin{equation}
            L_{peel} = \sum_{i=1}^{4} \Big\lVert {\widehat{ \mathcal{D}}_{peel}^i - \mathcal{D}_{peel}^i}\Big\lVert_1
        \end{equation}
        $L_{peel}$ captures the sum of $L_1$ norm between ground truth peeled depth map $\mathcal{D}_{peel}^i$ and predicted peeled depth map $\widehat{\mathcal{D}}_{peel}^i$ for each of the $i^{th}$ peeled map layer.
        \begin{equation}
            L_{rd} = \sum_{i=1}^{4} \Big\lVert{\widehat{ \mathcal{D}}_{rd}^i - \mathcal{D}_{rd}^i}\Big\lVert_1
        \end{equation}
        $L_{rd}$ constraints the RD predictions of SMPL to that of ground truth offsets. We also enforce per layer gradient smoothness of the predicted $(\widehat{\mathcal{D}}_{rd}^i+\mathcal{D}_{smpl}^i)$ and ground truth  $({\mathcal{D}}_{rd}^i+\mathcal{D}_{smpl}^i)$.
        \begin{equation}
                \begin{aligned}
                    L_{{sm}} = \sum_{i=1}^{4} \Big\lVert{\bigtriangledown ({\mathcal{D}}_{rd}^i+\mathcal{D}_{smpl}^i) - \bigtriangledown (\widehat{\mathcal{D}}_{rd}^i+\mathcal{D}_{smpl}^i)}\Big\Vert_1 
                \end{aligned}
            \label{eq:smoothloss}
        \end{equation}
Additionally, We also train our network with $L_1$ loss between predicted and ground truth RGB peeled maps ($L_{{rgb}}$).

   \subsection{Discussion on Architectural Choices}
    \label{subsubsec:arch-discuss}
    Existing prior-based non-parametric body reconstruction methods represent the body prior in 3D space (either voxels or meshes). However, this leads to computational inefficiencies as they need to densely probe the 3D space. Therefore, we adopted a sparse 2D peeled representation~\cite{jinka2020peeledhuman} of 3D shapes that can be efficiently processed by 2D convolution layers. Moreover, it enables to deform the body prior in order to approximate the loose clothing using residual deformation maps. Additionally, the peeled map representation also inherently addresses the self-occlusion issue. 
        \begin{figure}
        \includegraphics[width=\linewidth]{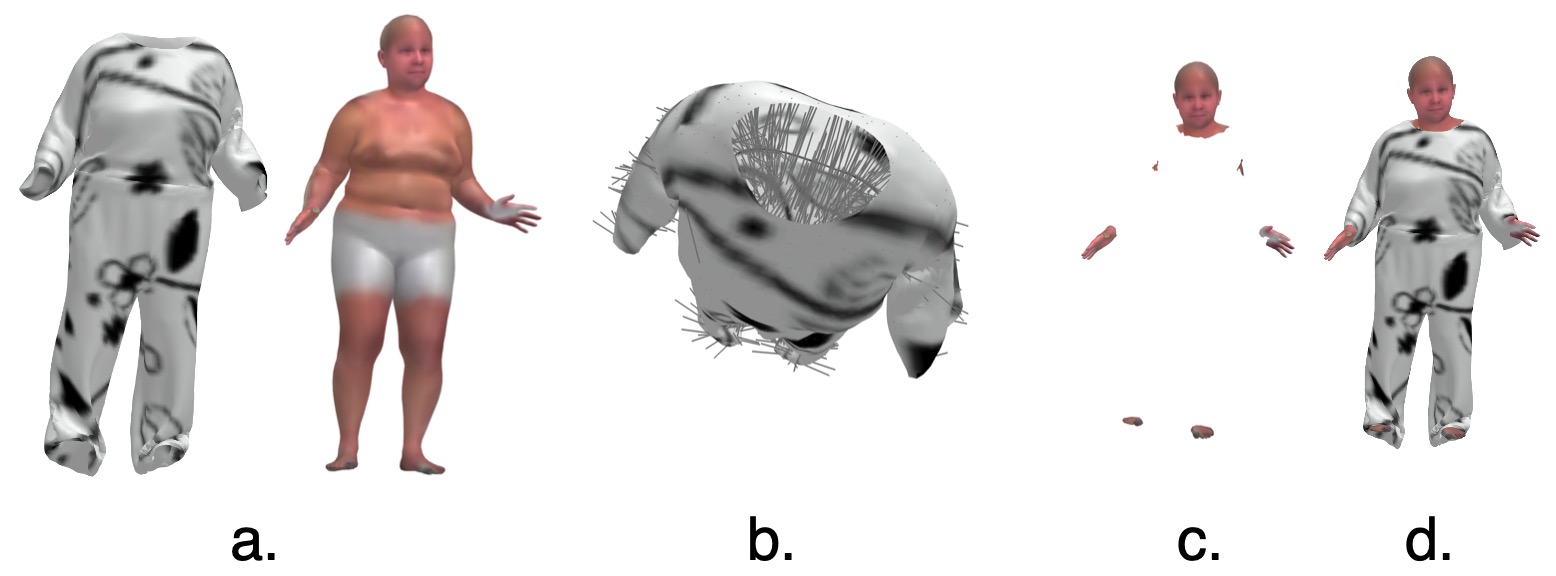}
        \caption{(a) Clothed mesh and SMPL body. (b) Casting rays inside clothing volume. (c) Faces of the body which intersect with the rays are removed. (d) final fused mesh.}
        \label{fig:dataset}
    \end{figure}

    In terms of network architecture, we use a shared encoder that encodes the peeled shape prior along with the input image using ResNet blocks into a shared latent representation. This enables a joint encoding of both geometry (coming from SMPL prior) and appearance and further regularizes the predicted peeled maps. 

    Regarding our peeled map fusion strategy, we model deformations of the SMPL prior surface as a pixel-wise residual offset in peeled layer representation. Unlike other existing methods (\cite{kolotouros2019convolutional, huang2020arch,zheng2019deephuman}) that model 3D deformations directly (in either volumetric or surface representations), our method is more efficient as we treat it as a simple pixel-wise offset. Additionally, this enables a more effective and natural fusion of predicted geometry of loose clothing with that of the underlying body. 
     %
        \begin{figure*}[bth]
        \centering
        \includegraphics[width=0.95\linewidth]{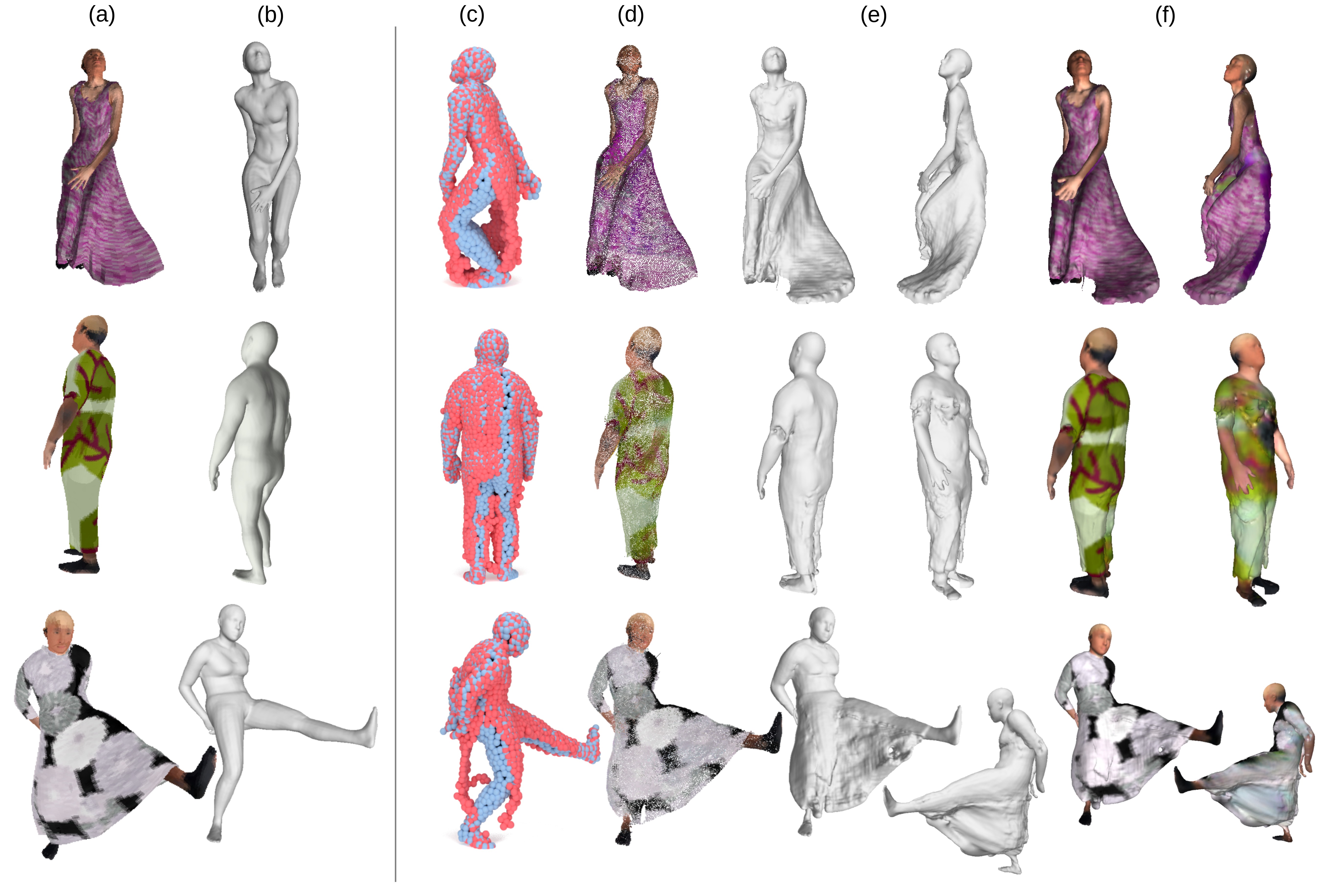} 
        \caption{\textbf{Qualitative results:} Given \textbf{(a)} an input image and \textbf{(b)} corresponding SMPL prior mesh, we render point-cloud of \textbf{(c)} predicted residual deformations (in red) added to the SMPL prior (in blue), \textbf{(d)} point-cloud obtained from the predicted peeled maps, and \textbf{(e)}, \textbf{(f)} the reconstructed mesh from two views. } 
    \label{fig:result_multiseqs}
    \end{figure*}
    \section{Experiments \& Results}
     \subsection{Implementation Details}
    Our multi-branch encoder-decoder network is trained end-to-end for $25$ epochs. The shared encoder consists of $3$ downsampling layers followed by $18$ ResNet blocks. Sigmoid activation is used in last layer of the $\widehat{\mathcal{D}}_{peel}$ and $\widehat{\mathcal{D}}_{rd}$ decoder branches while a tanh activation is used for the $\widehat{\mathcal{R}}$ decoder branch. The $\widehat{\mathcal{D}}_{rd}$ output values are scaled to a $[-0.15,0.15]$ range, which approximately maps to $[-15 \text{cm}, +15\text{cm}]$ metric scale range and holds empirical validity for cloth variations present in the datasets. Here, we use only $4$ layers of peeled representation, although the method is generalized to more number of peeled layers.

    We use the Adam optimizer with an exponentially decaying learning rate starting from $5 \times 10^{-4}$. Our network takes 30 hrs to train on $4$ Nvidia GTX $1080$ Ti GPUs with a batch size of $8$ and $\lambda_{rd}$, $\lambda_{rgb}$ and $\lambda_{sm}$ are set to $1, 0.1$ and $0.001$, respectively. We use~\cite{trimesh} for rendering the peeled maps.
    \subsection{Datasets}
        We perform both qualitative and quantitative evaluations on the following publicly datasets.
    \textbf{Cloth3D}~\cite{bertiche2020cloth3d} is a collection of $6500$ sequences of draped SMPL meshes simulated with MoCap data. Each frame of a sequence contains garment and body-specific deformation information. Garments styles range from skirts to very loose robes. We augment this data by capturing SMPL texture maps with minimal clothing to simulate realistic body textures using~\cite{alldieck19cvpr}. For each sequence, five frames are randomly sampled and the naked body is subtracted from the garment as shown in \autoref{fig:dataset} to obtain clothed mesh sans occluded body as ground truth for training. We use the ground truth SMPL parameter to get initial peeled maps prior for our training. We also augment this data by rotating each mesh by $45^\circ$, $60^\circ$, $-45^\circ$ along yaw axis to increase viewpoint variations.\\
    \\
    \textbf{THuman}~\cite{zheng2019deephuman} consists of 6800 human meshes registered with SMPL body in varying poses and garments. The dataset was obtained using consumer RGBD sensors. Although the dataset has diverse poses and shapes, it has relatively tight clothing examples with low-quality textures.
      \begin{table}
            \centering
            \begin{tabular}{lcc}
            \toprule
            Method & CD $\downarrow$ & P2S $\downarrow$\\
            \midrule
                PIFu~\cite{saito2019pifu} & 0.0064 &  0.0515\\
                ARCH~\cite{huang2020arch} & 0.0034 & 0.0357\\
                Geo-PIFu~\cite{he2020geo} & 0.0012 & 0.0256\\
                PeeledHuman~\cite{jinka2020peeledhuman}  & 0.0016 & 0.0291 \\
                \hline
                Ours (baseline) & 0.0010 &0.01417\\
                Ours & \textbf{0.0006} & \textbf{0.01204}\\
            \bottomrule
            \end{tabular}
            \caption{Quantitative comparison with state-of-the-art methods on Cloth3D data.}
        \label{table:comparisons}
    \end{table}
    \subsection{Qualitative Results}
\begin{figure*}
    \centering
        \includegraphics[width=0.95\linewidth]{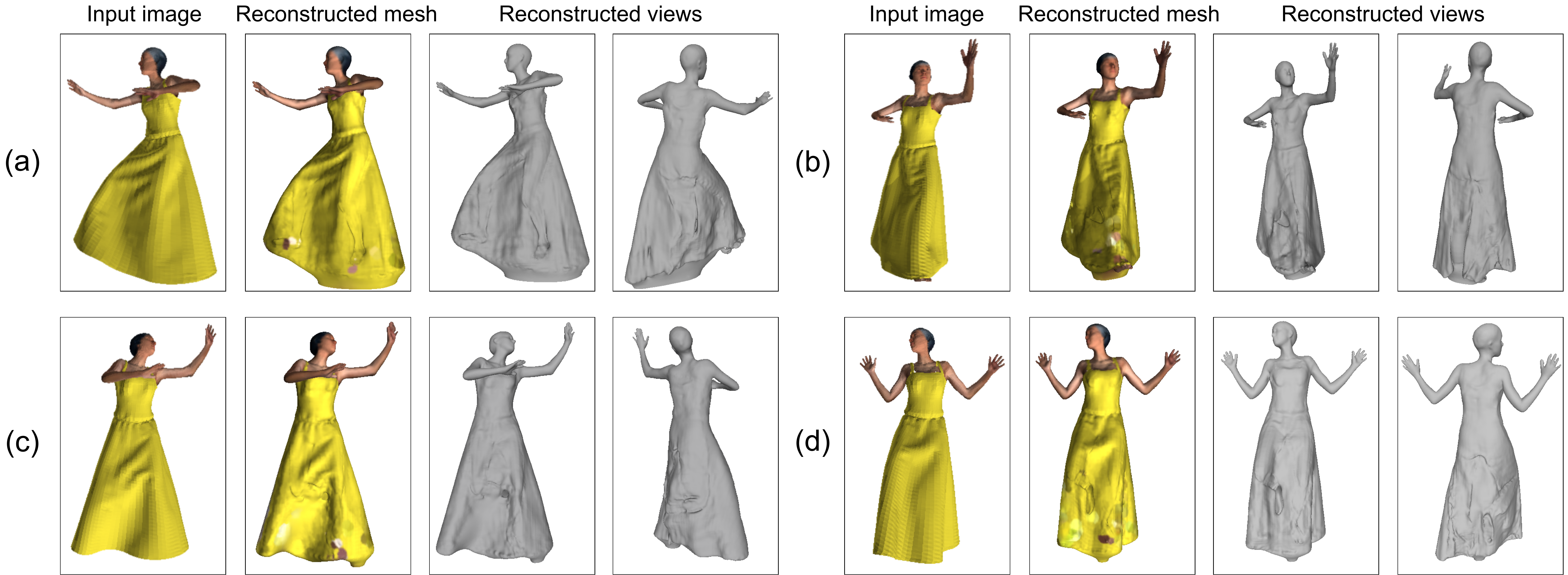}  
        \caption{\textbf{Qualitative results:} Our method predicts consistent body parts along with realistic clothing deformations from a monocular input image. Note that these are per-frame reconstructions. }
    \label{fig:result_singleseq}
    \end{figure*}

    Our method is able to predict high-fidelity human reconstructions with very loose clothing, as shown in \autoref{fig:result_multiseqs} and \autoref{fig:result_singleseq}. 
    We demonstrate qualitative results for different clothing styles in \autoref{fig:result_singleseq}, where we were able to successfully reconstruct complex body poses(bottom row). Please refer to supplementary material for additional results.
    \subsection{Quantitative Evaluation}
    To quantitatively evaluate performance of our method, we compute Point-to-Surface (P2S) distance and Chamfer Distance (CD) using the point-cloud obtained from the predicted fused maps. P2S calculates the distance between this point-cloud and the ground truth mesh. CD calculates the distance between this point-cloud and the point-cloud sampled from the ground truth mesh. 
    
    \subsection{Comparisons with the state-of-the-art}  
    \begin{table}
            \centering
            \begin{tabular}{lcc}
            \toprule
            Method & CD $\downarrow$ & P2S $\downarrow$\\
            \midrule
                DeepHuman~\cite{zheng2019deephuman} & 0.00119  & 0.00112\\
                Geo-PIFu~\cite{he2020geo} & 0.00017  &  \textbf{0.00019}\\
                \hline
                Ours & \textbf{0.00016} & \textbf{0.00019}\\
            \bottomrule
            \end{tabular}
            \caption{Quantitative comparisons on THuman dataset}
        \label{table:quant_tbl_THU}
    \end{table}
    We evaluate the aforementioned metrics on Cloth3D dataset on PIFu~\cite{saito2019pifu}, ARCH~\cite{huang2020arch}, GeoPIFu~\cite{he2020geo} and PeeledHuman~\cite{jinka2020peeledhuman}. Note that we retrained all these competing methods on Cloth3D data. We consider predicting peeled depth maps (before fusion) as our baseline. 
     \begin{figure}
    \includegraphics[width=\linewidth]{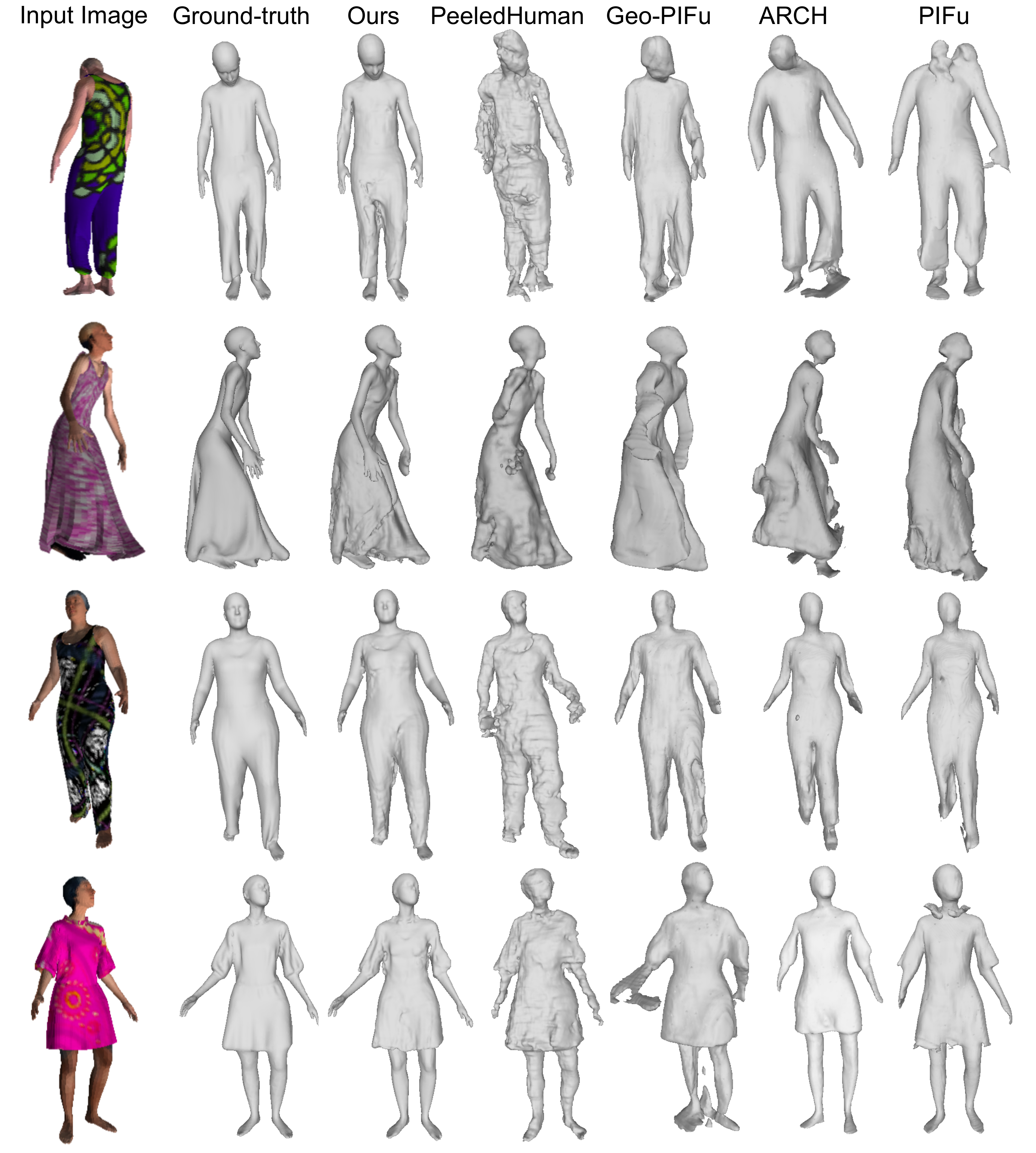}
    \caption{\textbf{Reconstruction Results (Cloth3D):} Our method yields superior geometrical reconstruction in comparison with PeeledHuman~\cite{jinka2020peeledhuman}, Geo-PIFu~\cite{he2020geo}, ARCH~\cite{huang2020arch} and PIFu~\cite{saito2019pifu} on Cloth3D dataset.} 
    \label{fig:comparison}
\end{figure}

     \begin{figure}
    \includegraphics[width=\linewidth]{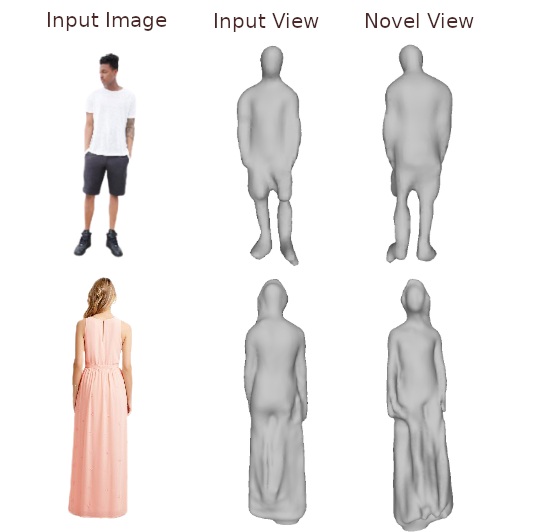}
    \caption{Inference results of our model on DeepFashion dataset, trained on Cloth3D dataset.}
    \label{fig:deepfashion}
\end{figure}

    Unlike~\cite{jinka2020peeledhuman} that uses GAN with CD loss, we use a simple encoder-decoder architecture without CD loss on the predicted depth maps. 
    To perform a fair comparison with ARCH, instead of sampling 3D points around the SMPL body in the canonical pose, we directly sample from the final pose for evaluation of labels. 
    Similar to other methods, we transform all the predicted models from different methods to the canonical coordinates of the ground truth mesh.

    \autoref{table:comparisons} summarizes the quantitative results where we outperform existing methods on Cloth3D data by a good margin for both CD and P2S metrics.  \autoref{fig:comparison} shows that our method consistently outperforms other existing methods in terms of quality of reconstructed geometrical details around both body and cloth region, over varying body shape and clothing styles. 
    
    We also evaluate our method on THuman dataset with the evaluation metric code provided by~\cite{he2020geo} and reported results in \autoref{table:quant_tbl_THU}. In this case, we perform on-par with GeoPIFu as THuman dataset has shapes with primarily tight clothing scenarios in contrast to Cloth3D dataset, which has widely loose clothing styles, where we outperform GeoPIFu~\cite{he2020geo}. Finally, we show reconstruction results on DeepFashion~\cite{liuLQWTcvpr16DeepFashion} dataset in \autoref{fig:deepfashion}.
    
    \begin{table}
    \centering
        \begin{tabular}{llll}
            \toprule
            & blocks & {Ours (baseline)} & {Ours}\\
            \midrule
            \multirow{3}{*}{CD} & {6} & {0.0013} & {0.0008} \\
                                & {9} & {0.00011} & {0.0006}\\
                                & {18} & {0.00010} & {0.0006}\\
            \midrule
            \multirow{3}{*}{P2S} & {6} & {0.01720} & {0.01452}\\
                                 & {9} & {0.0155} & {0.01338}\\
                                 & {18} & {0.01417} & {0.01204}\\
            \bottomrule
        \end{tabular}
       
    \caption{Effect of varying number of ResNet blocks.}
    \label{table:ablative_blks}
    \end{table}
\subsection{Ablation Study}
\begin{figure}
    \includegraphics[width=\linewidth]{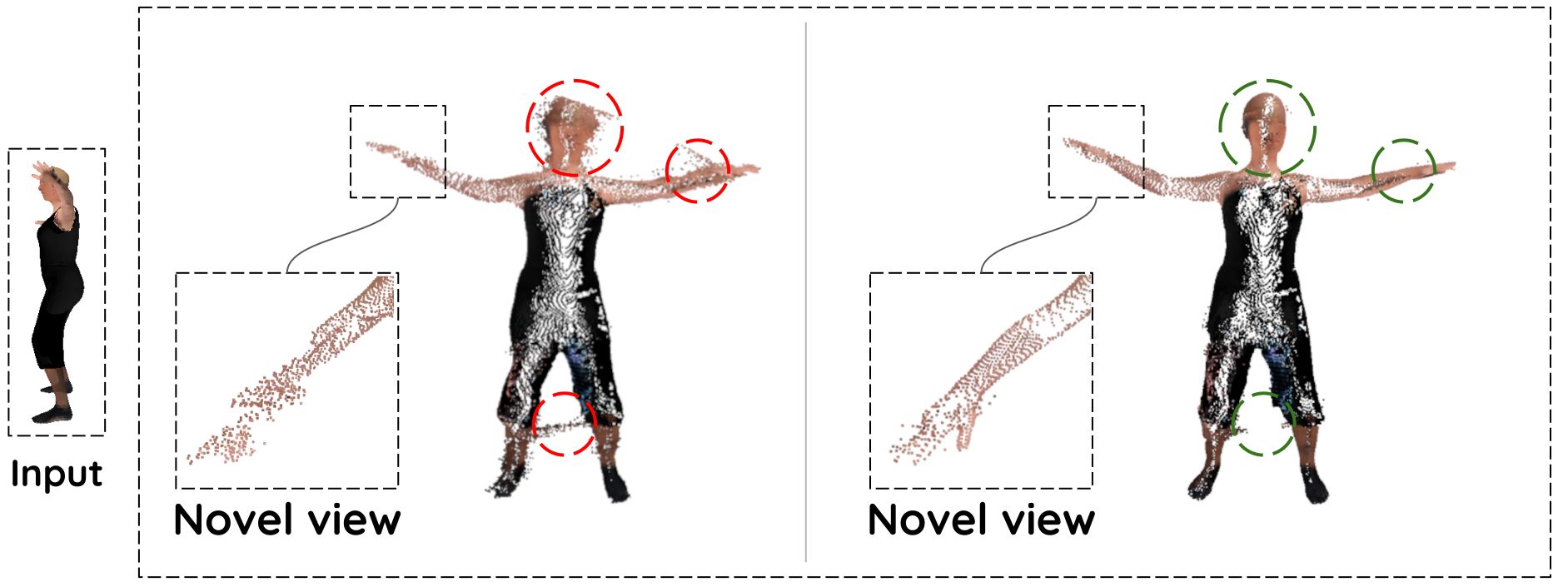}
    \centerline{\hspace{10mm}(a)\hspace{30mm}(b)}
    \newline
    \caption{Our peeled map fusion retains body parts in the final output (b) as compare to output of baseline setup (a).} 
    \label{fig:fusion}
\end{figure}

\textbf{Impact of Peeled Map Fusion:}
In \autoref{fig:fusion}, it can be observed that fusion of the RD maps with the predicted peeled maps helps to retain the body parts in complex and self-occluded poses. On the other hand, in the case of our baseline setup (without fusion module), we obtain distorted body parts. \\
\\
\textbf{Impact of Architecture Variations:}
We also evaluate the performance of our method by varying number of ResNet blocks in encoder as shown in \autoref{table:ablative_blks}.
We can observe that even with much smaller network size (6 ResNet blocks), our network is able to achieve better performance when compared to baseline setup and all other existing methods compared in \autoref{table:comparisons}.\\
\\
\begin{figure}
    \centering
    \includegraphics[width=0.95\linewidth]{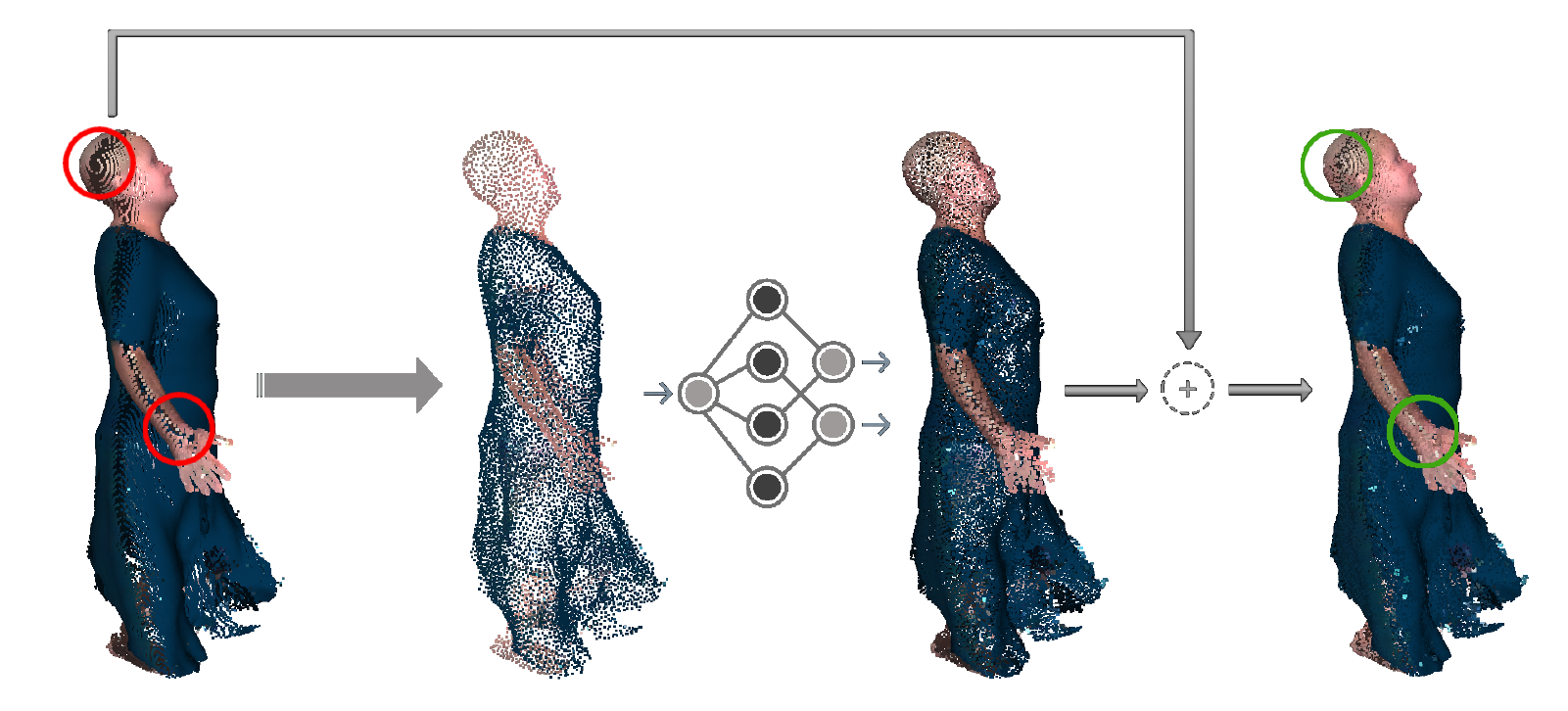}\newline
    \centerline{(a)\hspace{13mm}(b)\hspace{15mm}(c)\hspace{13mm}(d)\hspace{6mm}}
    \newline
    \caption{\textbf{Upsampling:} (a) Predicted  $\widehat{\mathcal{P}}$, (b) simplified $\widehat{\mathcal{P}}_{sub}$, (c) upsampled $\widehat{\mathcal{P}}_{dense}$ , (d) super-imposed PC.}
    \label{fig:upsampling}
    \end{figure}
    \\
\textbf{Effect of Upsampling on the Predicted Point-cloud} The predicted point-cloud $\widehat{\mathcal{P}}$, generated from the fused depth maps, can sometimes be sparse and contain holes. This issue arises because the perspective projection causes loss of information in the peeled maps, and this loss of information is reflected in the back-projected point-cloud.
 To tackle this issue, we create a simplified point-cloud $\widehat{\mathcal{P}}_{sub}$ by uniformly sampling $\sim20,000$ points from $\widehat{\mathcal{P}}$. Subsampling helps in avoiding the amplification of isolated noisy points, majorly from the parts that are not visible in the input image. Subsequently, we perform upsampling of $\widehat{\mathcal{P}}_{sub}$ using~\cite{qian2020pugeo} to obtain dense, upsampled point cloud $\widehat{\mathcal{P}}_{dense}$. The upsampling closes most of the potential holes but might lose the fine-grained geometrical details in $\widehat{\mathcal{P}}$. Hence, we superimpose $\widehat{\mathcal{P}}$  and $\widehat{\mathcal{P}}_{dense}$, which closes a significant number of holes and retains the geometrical details, as shown in ~\autoref{fig:upsampling}.


\section{Conclusion}
    We introduced a novel shape-aware peeled representation for the reconstruction of human bodies with loose clothing. Our method achieves sparse and efficient fusion of parametric body prior with non-parametric peeled depth representation.
    We evaluated our method on various publicly available datasets and reported superior qualitative and quantitative results as compared to state-of-the-art methods.
    As part of future work it would be interesting to introduce temporal consistency in the proposed solution. 

\bibliographystyle{ieee_fullname}
\bibliography{egbib}

\end{document}